\documentclass{article}

\PassOptionsToPackage{numbers,compress,square,sort,comma}{natbib}

\usepackage[preprint]{neurips_2019}

\usepackage[utf8]{inputenc} 
\usepackage[T1]{fontenc}    
\usepackage{hyperref}       
\usepackage{url}            
\usepackage{booktabs}       
\usepackage{amsfonts}       
\usepackage{nicefrac}       
\usepackage{microtype}      
\usepackage{graphicx}
\graphicspath{ {./imgs/} }
\usepackage[toc,page]{appendix}
\usepackage{caption}
\usepackage{tikz}
\usetikzlibrary{shapes,arrows}

\title{Transforming the output of GANs by fine-tuning them with features from different datasets}

\author{%
  Terence Broad \\
  Department of Computing\\
  Goldsmiths, University of London\\
  \texttt{t.broad@gold.ac.uk} \\
   \And
   Mick Grierson \\
   Creative Computing Institute \\
   University of the Arts London \\
   \texttt{m.grierson@arts.ac.uk} \\
}

\begin{document}

\maketitle

\begin{abstract}

In this work we present a method for fine-tuning pre-trained GANs with features from different datasets, resulting in the transformation of the output distribution into a new distribution with novel characteristics. The weights of the generator are updated using the weighted sum of the losses from a cross-dataset classifier and the frozen weights of the pre-trained discriminator. We discuss details of the technical implementation and share some of the visual results from this training process.

\end{abstract}

\section{Introduction}
The motivation for this work was to find a way of transforming a generative model that had been trained on one distribution, to output a completely new distribution of images that did not model an existing dataset. We approached this by taking the generator from a pre-trained generative adversarial network (GAN) \cite{goodfellow2014generative} trained on one dataset (in this case ImageNet \cite{deng2009imagenet}) and then fine-tuned it with features from another dataset using a classifier trained on data from both datasets.

With this approach we were hoping not to simply model the distribution of images in the new dataset, but transform the generator so it outputs a new distribution of images that fuses visual features from both datasets, resulting in a distribution with novel characteristics. By starting from a pre-trained model with good initial weights, we hoped that this would preserve some aspects of the original distribution, such as the spatial structure of the images, but instilling it with some new characteristics from the other dataset.

\section{Method}

We created a dataset of approximately 14k images from Pinterest boards with the title \mbox{\textit{a e s t h e t i c}}.\footnote{See Figure \ref{fig2} in the Appendix for samples.} Images from these boards can usually be characterised by having distinct, washed-our colour palettes  (often with only one dominant colour in the image) and often the photographs are framed with no particular subject in focus. 

We trained a binary classifier to classify between the \mbox{\textit{a e s t h e t i c}} images\footnote{We also trained classifiers for other datasets with prominent aesthetic characteristics, but for posterity, we will only be discussing results from fine-tuning with the classifiers trained on the \mbox{\textit{a e s t h e t i c}} dataset.} and images from the ImageNet dataset \cite{deng2009imagenet}. To train the classifier we fine-tuned a pre-trained ResNet \cite{he2016deep} model that had been trained to weakly classify Instagram hastags and then ImageNet \cite{mahajan2018exploring}. In addition to training the classifier to classify \mbox{\textit{a e s t h e t i c}} images and ImageNet images as separate classes (contrastive features), we also---initially by accident---trained a classifier that classifies them as being in the same class (joint features), which led to significantly better results when used for fine-tuning the generator (see Section \ref{discussion} for further discussion).

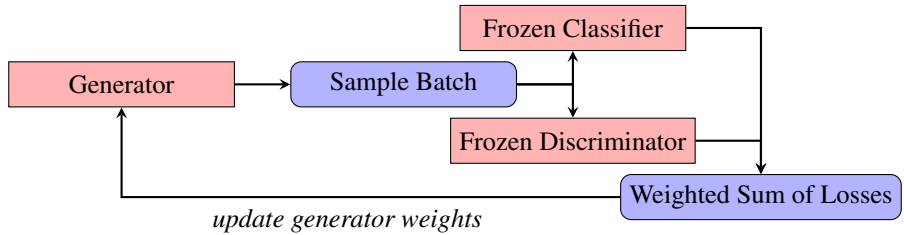
\begin{figure}
  \centering
\begin{tikzpicture}[auto, node distance=2cm,>=latex']

\tikzstyle{model} = [rectangle, minimum width=3cm, minimum height=0.6cm,text centered, draw=black, fill=red!30]
\tikzstyle{sample} = [rectangle, rounded corners, minimum width=3cm, minimum height=0.6cm,text centered, draw=black, fill=blue!30]
\tikzstyle{arrow} = [thick,->,>=stealth]

\node (generator) [model] {Generator};

\node (image) [sample] [right of=generator, xshift= 1.75cm] {Sample Batch};

\node (classifier) [model] [right of=generator, yshift= 0.75cm, xshift= 4cm] {Frozen Classifier};
\node (discriminator) [model] [right of=generator, yshift=-0.75cm, xshift= 4cm] {Frozen Discriminator};

\node (sum) [sample] [right of=generator, yshift=-1.5cm, xshift= 6.5cm] {Weighted Sum of Losses};

\draw [arrow] (generator) --(image);
\draw [arrow] (image) -| (classifier);
\draw [arrow] (image) -| (discriminator);
\draw [arrow] (classifier) -| (sum);
\draw [arrow] (discriminator) -| (sum);
\draw [arrow] (sum)  -| node[anchor=south, yshift=-0.6cm, xshift=3cm] {\textit{update generator weights}}(generator);
\end{tikzpicture}

  \caption{Diagram of training process: Batches of images are sampled from the pre-trained generator, which are fed to the cross-dataset classifier and the pre-trained discriminator (both of which have their weights frozen). The weights of the generator are updated based on a weighted sum of the losses from the classifier and discriminator.}
  \label{fig1}
\end{figure}

After training the cross-dataset classifier, we used this model to fine-tune the weights of a pre-trained BigGAN \cite{brock2018large} generator trained on the ImageNet dataset at a resolution of 128x128 pixels.\footnote{For this we used `The author's officially unofficial PyTorch BigGAN implementation' \url{https://github.com/ajbrock/BigGAN-PyTorch} and would like to thank the authors of the repository, Andrew Brock and Alex Andonian, for releasing the model weights for the discriminator as well as the generator, without which this work would not have been possible.} We also used the frozen weights of the discriminator in the fine-tuning training procedure, updating the weights of the generator based on a weighted sum of the loss from the discriminator and the cross-dataset classifier (see Figure \ref{fig1} for details). During this fine-tuning process, the networks are not exposed to any new training data, all the samples and losses are produced only using the pre-trained networks.

The process of training and convergence is very rapid. Usually within 1000 iterations (using a batch size of 9) the generator has converged onto a configuration of the weights that satisfies both the cross-dataset classifier and the discriminator. However we find that the best results were achieved using early stopping, often the most interesting visual results occurred when training was stopped after 300-600 iterations. Because training time is so quick, it is trivial to try multiple configurations of the parameter weighting and manually compare the visual results.

\section{Discussion and Conclusion}
\label{discussion}

In the process of this work we have happened upon a number of surprising results. The manner in which features get combined from the different datasets was highly unexpected. Neither the results of fine-tuning using the contrastive features or the joint features classifier have resulted in producing images that resemble the images in either the ImageNet or \mbox{\textit{a e s t h e t i c}} datasets. 

The second surprising result is that when fine-tuning with the joint features classifier the visual results were much richer and varied (almost dreamlike in nature) than the results from fine-tuning with the contrastive features classifier (see Figures \ref{fig4} and \ref{fig5} in the Appendix for a detailed comparison). We speculate that the contrastive features classifier discards a lot of important features from the ImageNet distribution, so when the generator is fine-tuned, there are less combinations of features that can be used and the resulting distribution has a lot less variety. 

In future research, we hope to find ways of having more control over what kind of characteristics from the different datasets get combined in the fine-tuning process, be that characteristics relating to aesthetic qualities, the structure and form in the images, or the stylistic qualities of a given dataset. We also hope to apply these techniques to higher resolution GAN models, but without having access to pre-trained discriminators, it is currently not possible to apply these techniques to the higher resolution generative models that have been made publicly available without retraining the models from scratch.

\subsubsection*{Acknowledgments}

This work has been supported by UK’s EPSRC Centre for Doctoral Training in Intelligent Games
and Game Intelligence (IGGI; grant EP/L015846/1).



\bibliographystyle{unsrt}
\bibliography{neurips_2019}

\appendixpage

\begin{figure}[!htb]
\centering
\captionsetup{justification=centering}
\includegraphics[width=0.5\textwidth]{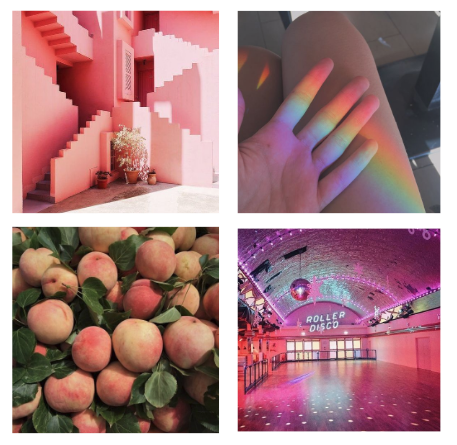}
\caption{Samples of images from the \mbox{\textit{a e s t h e t i c}} dataset sourced from Pinterest.}
\label{fig2}
\end{figure}

\begin{figure}[!htb]
\centering
\captionsetup{justification=centering}
\includegraphics[width=0.9\textwidth]{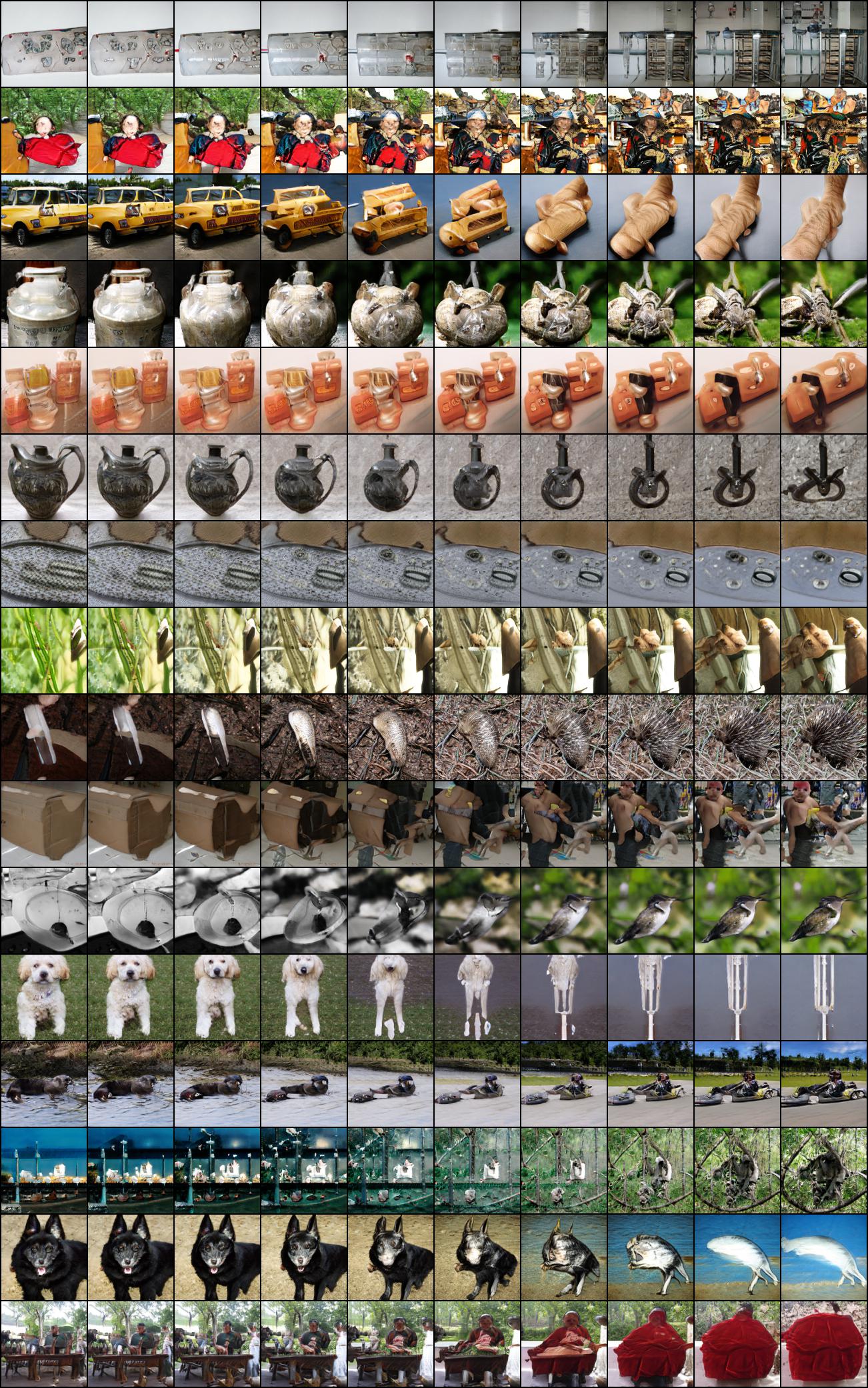}
\caption{Original BigGAN output (each row shows one interpolation between two classes). 
\break See Figures \ref{fig4} and \ref{fig5} for results of fine-tuning. }
\label{fig3}
\end{figure}

\begin{figure}[!htb]
\centering
\captionsetup{justification=centering}
\includegraphics[width=0.9\textwidth]{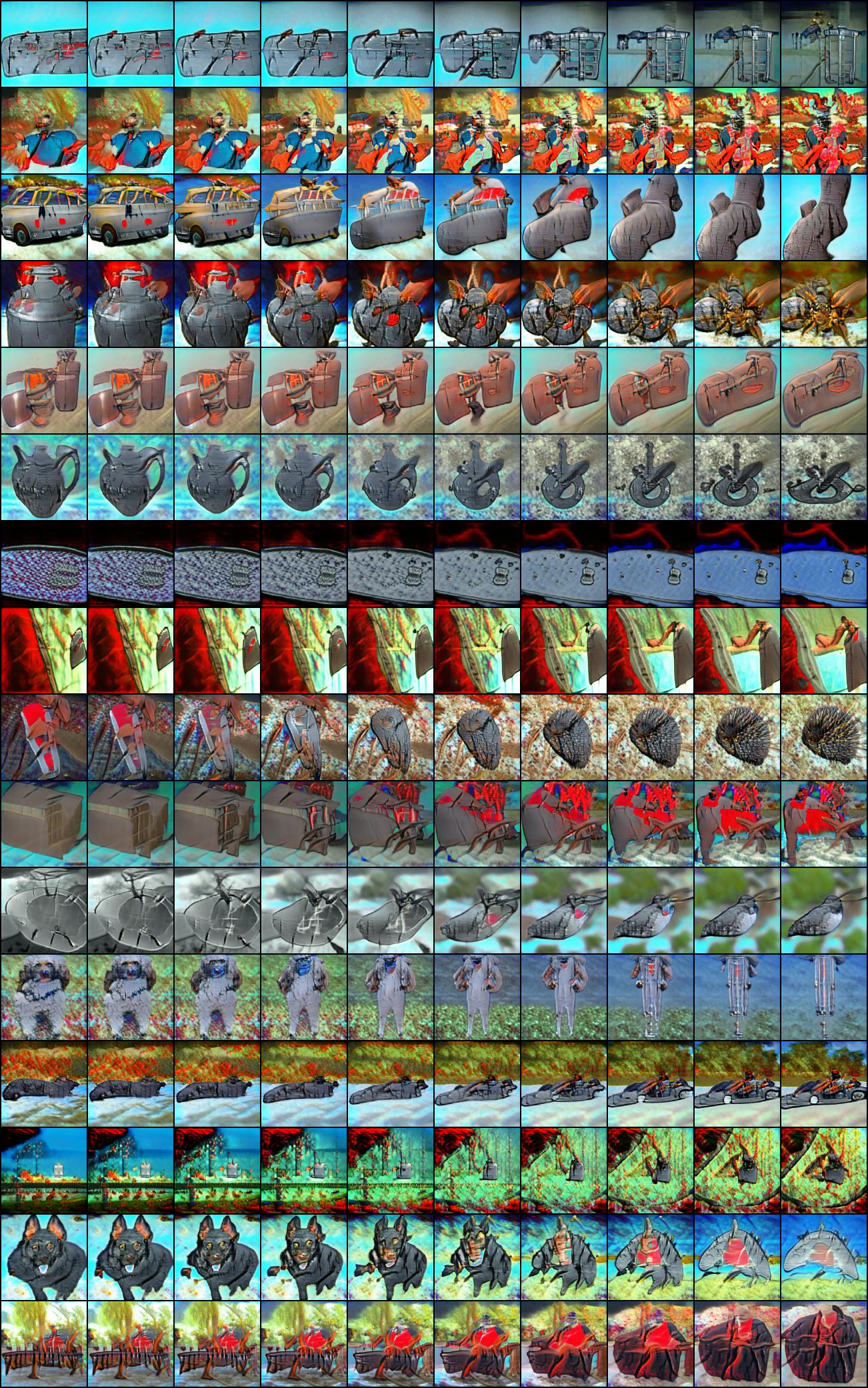}
\caption{Output of BigGAN fine-tuned with contrastive features classifier for 300 iterations. 
\break See Figure \ref{fig3} for reference of original BigGAN output.}
\label{fig4}
\end{figure}

\begin{figure}[!htb]
\centering
\captionsetup{justification=centering}
\includegraphics[width=0.9\textwidth]{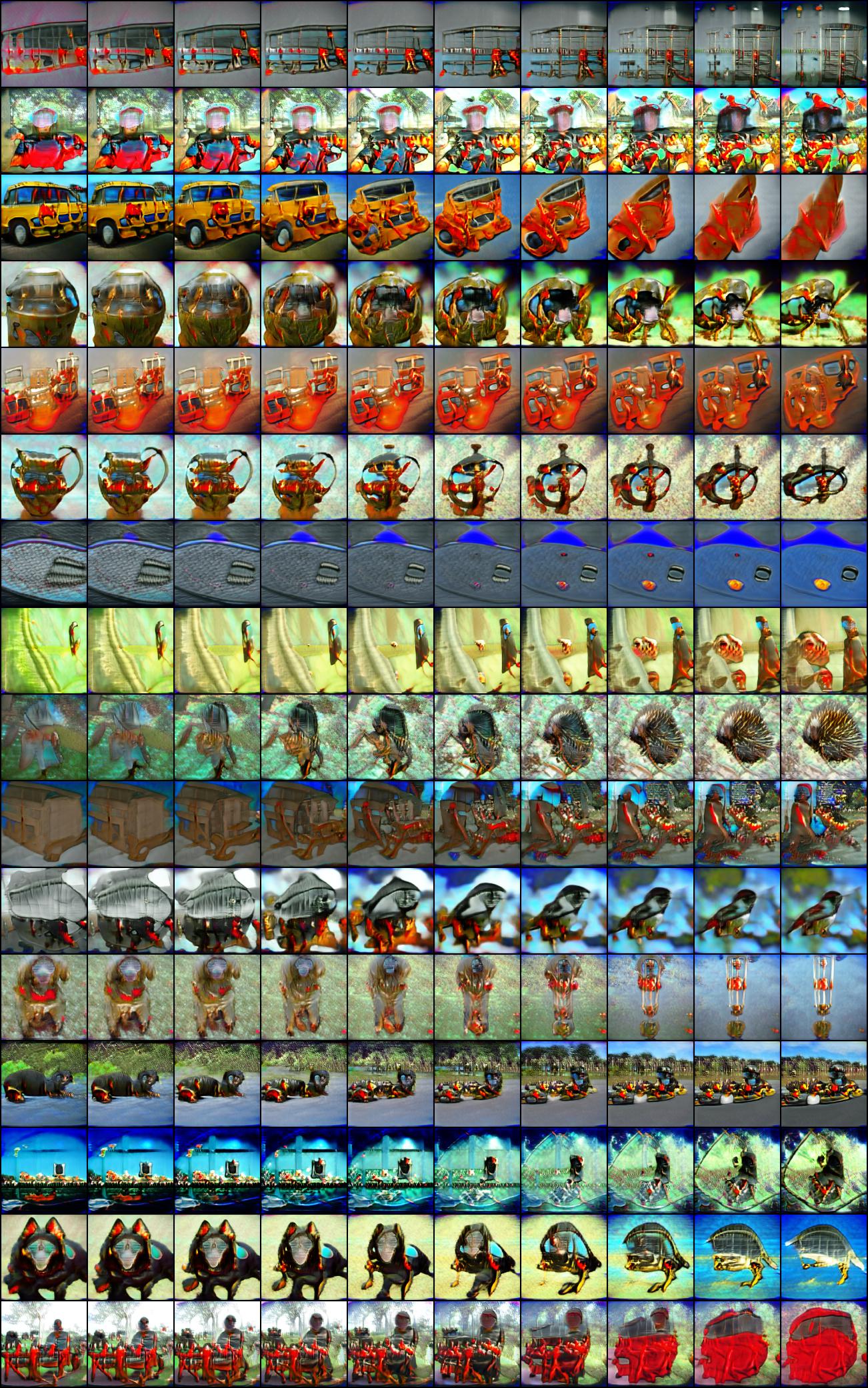}
\caption{Output of BigGAN fine-tuned with joint features classifier for 300 iterations. 
\break See Figure \ref{fig3} for reference of original BigGAN output.}
\label{fig5}
\end{figure}


\end{document}